\documentclass{article}

\usepackage{iclr2026_conference}
\usepackage{times}
\usepackage{url}
\usepackage{graphicx}
\usepackage{caption}
\usepackage{subcaption}
\usepackage{amsmath}
\usepackage{amssymb}
\usepackage{booktabs}

\usepackage{algorithm}
\usepackage{algorithmic}

\usepackage{newfloat}
\usepackage{listings}
\DeclareCaptionStyle{ruled}{labelfont=normalfont,labelsep=colon,strut=off}
\floatstyle{ruled}
\newfloat{listing}{tb}{lst}{}
\floatname{listing}{Listing}

\title{Pruning via Causal Attribution Preserves \\ Reasoning Performance in Large Language Models}

\iclrfinalcopy

\author{Amogh Sheth \\
Edison Academy Magnet School \\
\texttt{amogh.sheth1@gmail.com} \\
\And
Andrew Lin \\
Massachusetts Institute of Technology \\
\texttt{a\_lin@mit.edu} \\
\AND
Biruk Assefa \\
State University of New York College at Plattsburgh \\
\texttt{basse002@plattsburgh.edu} \\
\AND
Yi Wen Huang \\
The University of Texas at Austin \\
\texttt{yiwenhuang@utexas.edu} \\
\And
Yuhao Ge \\
Independent Researcher \\
\texttt{geyuhao33@gmail.com}
}

\begin{document}
\maketitle

\begin{abstract}
Large language models (LLMs) excel at multi-step reasoning but incur substantial inference cost.
We introduce \emph{Causal Attribution Pruning} (CAP), a training-free method that identifies critical \emph{attention heads} by measuring their causal impact on reasoning tasks and uses these head-level scores to guide fine-grained weight pruning.
For each attention head, CAP estimates the expected performance degradation when the head is masked during forward passes on a small calibration set of reasoning problems.
These causal scores are then converted into weight-level importance values for the corresponding projection matrices.
Unlike magnitude-only or activation-based criteria, CAP's interventional measurement directly captures each head's functional contribution, yielding relative accuracy gains of up to 61\% over Wanda on ARC-Challenge at 20\% sparsity.
We evaluate CAP on GSM8K, StrategyQA, and ARC-Challenge using Llama-3-8B-Instruct and Mistral-7B-Instruct at 10\%, 20\%, and 50\% sparsity.
At moderate sparsity (10--20\%), CAP improves over Wanda in most model--benchmark configurations, with especially large gains on ARC-Challenge for Llama-3.
Our results suggest that attention-head-level causal attribution can better preserve reasoning performance on downstream benchmarks than correlational pruning criteria at equivalent sparsity, while remaining limited by coarse MLP attribution at 50\% sparsity.
\end{abstract}

\section{Introduction}

Scaling has driven most advances in modern AI, but at the cost of substantially increasing compute and latency, creating barriers to real-world deployment.
Model pruning addresses this by removing redundant parameters---such as weights within attention heads or MLP layers---to reduce size and inference cost without retraining.
While effective for compression, existing pruning methods can substantially degrade performance on complex tasks that require multi-step reasoning.

Most pruning approaches rely on \emph{correlational} signals such as weight magnitude or local gradient statistics.
Magnitude-based pruning assumes that small weights are unimportant, but even low-magnitude parameters can play essential roles in multi-step reasoning.
Gradient-based methods, though more adaptive, capture only local sensitivity and can overlook parameters that become crucial when reasoning across contexts.
These limitations frequently cause \emph{layer collapse}, where entire modules lose function, leading to sharp degradation on reasoning-heavy benchmarks \citep[e.g.][]{frantar2023sparsegptmassivelanguagemodels,sun2024simpleeffectivepruningapproach}.

To address these issues, we introduce \textbf{Causal Attribution Pruning (CAP)}, a training-free method that measures the \emph{causal contribution} of each attention head to reasoning performance and uses these measurements to guide weight-level pruning at inference time.
By estimating the expected performance drop when individual heads are masked on a small calibration set, CAP moves beyond correlational proxies to quantify which heads are functionally critical for downstream tasks.
These head-level scores are then propagated to individual weights within the corresponding projection matrices, yielding a pruning criterion that is both interventionally grounded and parameter-specific.

\paragraph{Scope of causal attribution.}
We emphasize that CAP's causal scoring operates at the \emph{attention head} level: each head is individually masked and its impact on task loss is measured.
MLP layers, which lack the same natural head-partitioned structure, receive an approximate per-layer importance derived from their co-located attention heads (Section~\ref{sec:weight_pruning}).
We do not claim fine-grained causal attribution over individual MLP neurons or other sub-components; extending causal interventions to these modules is an important direction for future work.

\paragraph{Why focus on chain-of-thought reasoning?}
Chain-of-thought (CoT) reasoning is a cornerstone capability of modern instruction-tuned LLMs \citep{Wei2022ChainOfThought}, enabling them to solve mathematical word problems, multi-hop question answering, and scientific reasoning.
Preserving this capability under pruning is critical for deploying compressed models in real-world applications.
CAP is therefore calibrated on reasoning data so that causal attribution scores align with the target capability, rather than with generic language modeling behavior.

\noindent\textbf{Contributions.}
\begin{itemize}
    \item We formalize \emph{attention-head-level causal attribution} for pruning, measuring the expected increase in task-specific loss when masking an individual attention head relative to the intact model.
    \item We introduce a budget-aware procedure that converts these head-level causal scores into weight-level importance values, enabling fine-grained magnitude pruning under a global sparsity constraint while preserving head structure.
    \item We evaluate CAP on GSM8K, StrategyQA, and ARC-Challenge using Llama-3-8B-Instruct and Mistral-7B-Instruct, showing improvements over Wanda in most 10--20\% sparsity settings, with relative gains of up to 61\% on ARC-Challenge.
\end{itemize}

\section{Related Work}

\paragraph{Magnitude- and activation-based pruning.}
Early pruning methods removed parameters with small absolute values, demonstrating that dense networks can be substantially compressed without retraining \citep{han2015learningweightsconnectionsefficient,han2016deepcompressioncompressingdeep}.
Recent training-free methods for LLMs extend this paradigm by ranking weights using correlational statistics and performing one-shot pruning at scale.

SparseGPT \citep{frantar2023sparsegptmassivelanguagemodels} uses second-order (Hessian) information and applies compensatory weight updates during pruning, achieving strong perplexity preservation even at high sparsity; however, its weight-update step makes it more computationally expensive than simpler score-and-threshold approaches.
Wanda \citep{sun2024simpleeffectivepruningapproach} offers a lighter alternative that ranks weights by the product of magnitude and input activation norm, achieving competitive results without weight updates.
Both methods perform well on perplexity and standard downstream tasks, but our results indicate that such correlational proxies can be unreliable for multi-step reasoning: low-magnitude parameters may nonetheless serve as crucial connections in reasoning circuits, and locally defined criteria do not capture cross-layer interactions that matter for chain-of-thought inference.

\paragraph{Attribution and causal analyses.}
A complementary line of work employs explainability signals to guide compression.
In computer vision, ``pruning by explaining'' uses Layer-wise Relevance Propagation or Integrated Gradients to assign relevance scores to channels, improving the accuracy--compression trade-off for CNNs and vision Transformers \citep{Hatefi2024PruningByExplaining,Yu2023XPruner}.
For language models, attribution-guided pruning based on integrated gradients has been explored for tasks such as translation and summarization \citep{AttributionPruningLLMs2025}.
However, gradient-based attributions capture correlational sensitivity that can be noisy and highly input-local; large gradients do not necessarily imply that ablating a component will harm performance, nor do small gradients guarantee redundancy.
In parallel, causal studies such as Causal Head Gating \citep{Nam2025CHG} intervene directly on attention heads to characterize their functional roles by measuring the behavioral impact of head masking.
These works provide insight into the internal organization of Transformer models but generally do not define budget-aware pruning rules and are seldom evaluated on chain-of-thought reasoning.

\paragraph{Positioning.}
CAP lies at the intersection of these directions.
Like magnitude-based approaches, it produces unstructured weight-level sparsity patterns compatible with standard LLM inference infrastructure.
Unlike purely correlational criteria, CAP derives \emph{attention-head} importance scores from explicit masking interventions on a reasoning-focused calibration set, then propagates these scores to guide weight-level magnitude pruning under a global sparsity budget.
We compare primarily against Wanda, our like-for-like training-free baseline.
We revisit the absence of broader baselines such as SparseGPT in Section~\ref{sec:limitations}.

\section{Method}

\begin{figure*}[t]
\centering
\includegraphics[width=\textwidth,height=0.28\textheight,keepaspectratio]{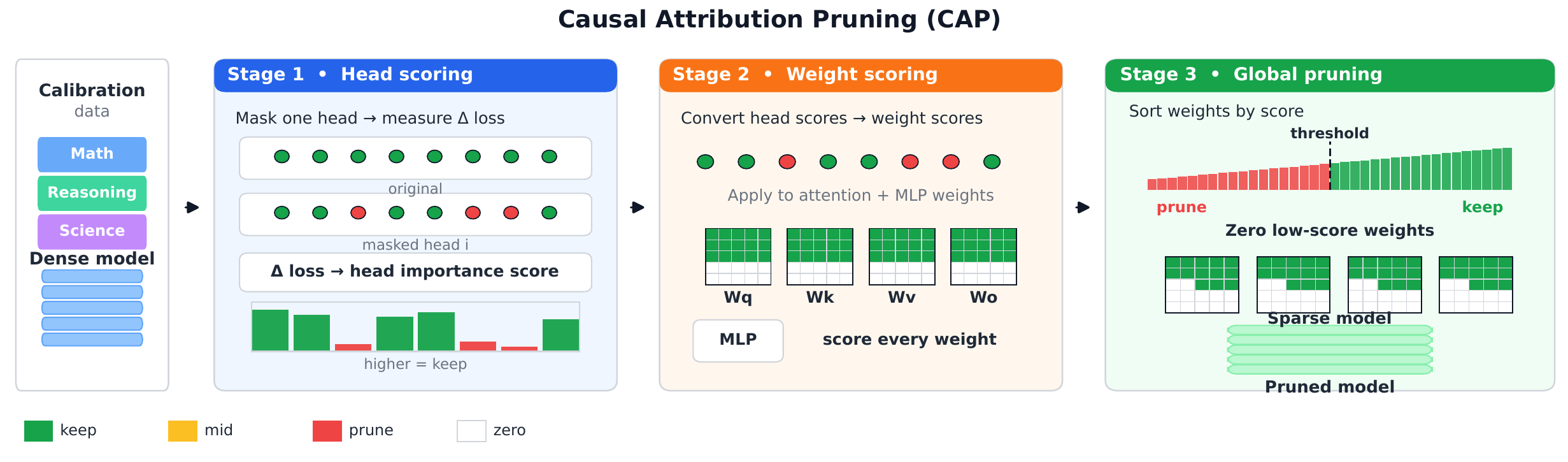}
\caption{\textbf{Overview of the CAP framework.} \textbf{Stage~1 (blue):} Mask each attention head and measure the loss increase $\Delta(h_\ell)$; high-$\Delta$ heads are critical (green) and low-$\Delta$ heads are redundant (red). \textbf{Stage~2 (orange):} Convert head-level causal scores into per-weight importance factors that scale weight magnitudes for attention projections ($W_q/W_k/W_v/W_o$) and MLP weights. \textbf{Stage~3 (green):} Globally rank all weights by importance-weighted magnitude and prune to the target sparsity.}
\label{fig:method_overview}
\end{figure*}

We introduce \emph{Causal Attribution Pruning} (CAP), a training-free method that identifies critical attention heads in large language models by measuring their interventional contribution to reasoning performance.
CAP measures \emph{causal} importance via head masking, then uses these scores to guide \emph{fine-grained magnitude pruning at the weight level}, zeroing out non-essential parameters within head-specific regions at inference time.
Figure~\ref{fig:method_overview} illustrates the three-stage framework.

\subsection{Calibration Dataset Construction}

We construct a calibration dataset $\mathcal{D}_{\text{cal}}$ by sampling from three reasoning-centric benchmarks: GSM8K \citep{Cobbe2021GSM8K} (grade-school math word problems), StrategyQA \citep{Geva2021StrategyQA} (multi-hop commonsense reasoning), and ARC-Challenge \citep{Clark2018ARC} (science questions requiring world knowledge and inference).
We sample $n\in\{128, 256\}$ examples per task (disjoint from the evaluation set), yielding a total of $|\mathcal{D}_{\text{cal}}| = 384$--$768$ reasoning-rich input--target pairs $(x, y)$.
We use stratified sampling across length-based quartiles as a rough proxy for problem complexity.
All sampling is performed with fixed random seeds (seed=0) to ensure reproducibility.

\subsection{Causal Attribution Score}

For an attention head $h$ in layer $\ell$, we define its causal attribution score as the expected increase in token-level cross-entropy when $h$ is temporarily masked during inference:

\begin{equation}
\Delta(h_\ell) \;=\; \mathbb{E}_{(x,y)\sim \mathcal{D}_{\text{cal}}}\!\left[ \,\mathcal{L}\!\left(f^{\text{mask}(h_\ell)}(x), y\right) \;-\; \mathcal{L}\!\left(f(x), y\right) \right],
\end{equation}

where $f$ is the intact model, $f^{\text{mask}(h_\ell)}$ is the model with head $h$ masked, $\mathcal{L}$ denotes cross-entropy loss, and $\mathcal{D}_{\text{cal}}$ is the calibration set. A higher $\Delta(h_\ell)$ indicates that masking head $h$ significantly degrades model performance, signaling its causal importance for the task.

\paragraph{Temporary head masking for scoring.}
We implement masking with forward hooks that intervene \emph{after head aggregation and before the output projection}.
This ensures that we measure the contribution of each head's \emph{output} to the final prediction, rather than disrupting internal query--key--value computations (which would introduce confounds from altered attention patterns).
Specifically, we apply the mask to the attention output tensor $\mathbf{X}\in\mathbb{R}^{B\times T\times H\times D}$, where $B$ is batch size, $T$ is sequence length, $H$ is the number of heads, and $D$ is the head dimension.
With a binary head mask $\mathbf{m}\in\{0,1\}^H$, the masked output zeros out the contribution of the target head while leaving the output projection $W_o$ intact during scoring.

\paragraph{Robust estimation via median aggregation.}
To reduce variance inherent in small-sample causal estimation, we partition $\mathcal{D}_{\text{cal}}$ into $K\in\{3,5\}$ disjoint subsets and compute $K$ independent estimates. We then take the median and subtract a multi-run intact baseline:

\begin{equation}
\widehat{\Delta}(h_\ell) \;=\; \mathrm{median}\!\left\{\Delta_k(h_\ell)\right\}_{k=1}^{K} \;-\; \widehat{\Delta}_{\text{base}},
\end{equation}

where $\widehat{\Delta}_{\text{base}}$ is the mean intact loss computed over the same $K$ subsamples.
We use the median for masked estimates because head ablations occasionally induce large loss spikes on a few examples, whereas the intact baseline is much more stable and is therefore summarized by its mean.
This mixed aggregation is robust to outliers while preserving an unbiased reference point for intact-model performance.
Our analyses in Section~\ref{sec:ablation} suggest that without this technique, head importance rankings become less stable, leading to worse pruning decisions.

\paragraph{Highly variable importance distribution.}
We observe striking variability in head utility across models.
In Llama-3, a small subset of heads (notably in layers 11--14 and at layer boundaries 0--2, 30--31) exhibit dramatically higher $|\Delta(h_\ell)|$ values, indicating their removal would catastrophically degrade reasoning.
The majority of heads in middle layers (15--27) contribute minimally, with near-zero causal scores.
Mistral exhibits a similar but more uniform pattern, with critical heads concentrated in early layers (0--5) and a flatter distribution in deeper layers.
This variability directly motivates our weight-level approach: uniform layer-wise or head-wise pruning strategies would either eliminate critical reasoning circuits or fail to exploit the abundant redundancy in low-impact regions.
We visualize these importance patterns in Figure~\ref{fig:head_heatmap} (Appendix~\ref{app:head_heatmaps}).

\subsection{Weight-Importance Guided Pruning}
\label{sec:weight_pruning}

Rather than removing entire heads (which can break grouped-query attention or key--value sharing in architectures like Llama-3 and Mistral), CAP uses the head-level causal scores to guide fine-grained, weight-level magnitude pruning.
The core idea is to combine two sources of information: (1)~\emph{causal head scores} that identify which attention heads are critical for reasoning, and (2)~\emph{weight magnitudes} that indicate which individual parameters within those heads have low numerical salience.
By multiplying these factors, we create a pruning metric that is both causally informed and parameter-specific.

\paragraph{Importance factor conversion.}
We convert the empirical causal scores into normalized importance factors using a shift-and-scale transformation:

\begin{equation}
I(h_\ell) \;=\; 1.0 + \frac{(\widehat{\Delta}(h_\ell) - \Delta_{\min}) \cdot 9.0}{\Delta_{\max} - \Delta_{\min}},
\end{equation}

where $\Delta_{\min}$ and $\Delta_{\max}$ are the minimum and maximum causal scores across all heads.
This maps scores to a bounded range $[1.0, 10.0]$ while preserving relative ordering.
We choose this 10$\times$ dynamic range as a simple heuristic that provides enough contrast to protect high-impact heads without overwhelming the underlying magnitude information.
A head with high causal impact receives $I(h_\ell) \approx 10.0$, protecting its weights from pruning; heads with minimal impact receive $I(h_\ell) \approx 1.0$, marking their weights as candidates for removal.
When all heads have identical scores, we assign uniform importance $I(h_\ell) = 1.0$.

\paragraph{Weight-level scoring.}
For each linear layer (attention projections $\mathbf{W}_q,\mathbf{W}_k,\mathbf{W}_v,\mathbf{W}_o$ and MLP layers $\mathbf{W}_{\text{gate}},\mathbf{W}_{\text{up}},\mathbf{W}_{\text{down}}$), we assign each weight $W_{ij}$ to a head region and compute:

\begin{equation}
S_{ij} \;=\; |W_{ij}| \cdot I\!\big(h_{\mathrm{region}(i,j)}\big).
\end{equation}

Weights in high-importance heads get amplified scores and are less likely to be pruned, while weights in low-importance heads retain scores close to their raw magnitude.

\paragraph{Region mapping.}
The region mapping follows the head partitioning of the weight matrices:

\begin{itemize}
    \item For $\mathbf{W}_o\!\in\!\mathbb{R}^{(H \cdot d_h)\times d_{\text{model}}}$, \emph{columns} are partitioned by head: $h = \lfloor j / d_h \rfloor$.
    \item For $\mathbf{W}_q,\mathbf{W}_k,\mathbf{W}_v\!\in\!\mathbb{R}^{d_{\text{model}}\times(H \cdot d_h)}$, \emph{rows} are partitioned by head: $h = \lfloor i / d_h \rfloor$.
    \item For MLP layers, we assign a per-layer average importance computed from all attention heads in that layer. This is an \emph{approximation}: it does not capture the fine-grained importance of individual MLP neurons, and it assumes that MLP importance covaries with attention head importance within a layer. We discuss this limitation in Section~\ref{sec:limitations}.
\end{itemize}

\emph{Weight-level pruning} refers to the fact that we prune individual scalar parameters $W_{ij}$ rather than entire rows, columns, or head blocks, enabling unstructured sparsity patterns that adapt to the learned weight distributions.

\paragraph{Global threshold selection.}
We choose a global threshold $\tau$ such that pruning all weights $W_{ij}$ with $S_{ij}\le \tau$ meets the target sparsity budget.
We employ an iterative binary search over $\tau$ to hit the target sparsity within $\pm 0.01\%$.
This global rule prioritizes removal from low-causal-utility regions while preserving fine-grained control.
At very high sparsity (e.g., 50\%), even weights within causally critical heads may fall below the threshold, leading to the removal of parameters essential for basic functionality.

\subsection{Implementation Details}

\paragraph{Calibration data.} We target CoT-style reasoning by sampling 128--256 examples per task from GSM8K, StrategyQA, and ARC-Challenge (disjoint from evaluation splits), partitioning them into 3--5 disjoint subsamples for robust scoring, and using fixed seeds for reproducibility. We also report perplexity on WikiText-2 \citep{Merity2017WikiText} (100 samples, ${\sim}$204k tokens) to assess generalization beyond reasoning tasks.

\paragraph{Dynamic masking.} Forward hooks apply binary masks at the attention output tensor $(B\times T\times H\times D)$, leaving weights unchanged during scoring. This enables exact, reversible interventions during scoring without modifying the underlying parameters.

\paragraph{Weight-level execution.} After scoring, we prune by setting selected weights to exactly zero (in-place) and save pruned checkpoints using safe serialization with 4\,GB shards for compatibility with Hugging Face Transformers \citep{Wolf2020Transformers}.

\paragraph{Numerical stability.} We guard attention masks and position IDs to prevent NaN propagation, and refine $\tau$ iteratively via binary search to hit the target sparsity within $\pm 0.01\%$.

\paragraph{Caching.} Head scores are cached per model and calibration configuration (keyed by model ID, dataset, and sample count) to enable multi-sparsity exploration without re-scoring. Cache hits reduce pruning time by $3$--$5\times$.

Algorithm~\ref{alg:cap} summarizes the full procedure.

\begin{algorithm}[t]
\caption{Causal Attribution Pruning (CAP)}
\label{alg:cap}
\begin{algorithmic}[1]
\STATE \textbf{Input:} model $f$, calibration set $\mathcal{D}_{\text{cal}}$, target sparsity $s$, subsamples $K$
\STATE Partition $\mathcal{D}_{\text{cal}}$ into $K$ disjoint subsets; estimate $\widehat{\Delta}_{\text{base}}$ from intact runs
\STATE \textbf{// Stage 1: Causal attention head scoring}
\FOR{each attention head $h_\ell$ in all layers}
  \FOR{$k=1$ to $K$}
    \STATE Attach forward hook to mask $h_\ell$ at attention outputs; compute $\Delta_k(h_\ell)$ on subset $k$
  \ENDFOR
  \STATE $\widehat{\Delta}(h_\ell) \gets \mathrm{median}\{\Delta_k(h_\ell)\}_{k=1}^K - \widehat{\Delta}_{\text{base}}$
\ENDFOR
\STATE \textbf{// Stage 2: Normalize head importance scores}
\STATE $\Delta_{\min} \gets \min_{\ell,h} \widehat{\Delta}(h_\ell)$, $\Delta_{\max} \gets \max_{\ell,h} \widehat{\Delta}(h_\ell)$
\FOR{each attention head $h_\ell$}
  \STATE $I(h_\ell) \gets 1.0 + \frac{(\widehat{\Delta}(h_\ell) - \Delta_{\min}) \cdot 9.0}{\Delta_{\max} - \Delta_{\min}}$
\ENDFOR
\STATE \textbf{// Stage 3: Weight-level scoring and pruning}
\FOR{each linear layer (attn projections + MLPs)}
  \STATE Map weights $W_{ij}$ to head regions; compute $S_{ij}=|W_{ij}|\cdot I(h_{\mathrm{region}(i,j)})$
\ENDFOR
\STATE Sort all weights ascending by $S_{ij}$; find threshold $\tau$ for target sparsity $s$
\STATE Set weights with $S_{ij} \leq \tau$ to zero; save checkpoint
\STATE \textbf{return} pruned model $f_{\text{pruned}}$
\end{algorithmic}
\end{algorithm}

\section{Experimental Setup}

We evaluate CAP on instruction-tuned large language models.
Llama-3-8B-Instruct \citep{Touvron2023Llama} and Mistral-7B-Instruct-v0.2 \citep{Jiang2023Mistral} are widely adopted open-weight instruction-tuned models, both employing grouped-query attention (GQA), which makes na\"ive head-level structural pruning problematic because heads share key--value projections.
We also conduct exploratory experiments on a mixture-of-experts model to assess generalization to architectures with dynamic routing.

Our evaluation focuses on three reasoning-centric benchmarks.
GSM8K \citep{Cobbe2021GSM8K} contains grade-school math word problems requiring arithmetic reasoning and step-by-step decomposition.
StrategyQA \citep{Geva2021StrategyQA} tests multi-hop commonsense reasoning where models must combine world knowledge across multiple inference steps.
ARC-Challenge \citep{Clark2018ARC} presents science questions requiring domain knowledge and logical reasoning.
For GSM8K, we report exact match on the final numerical answer.
StrategyQA is evaluated using binary accuracy, and ARC-Challenge by multiple-choice accuracy.

To align pruning with the target capability, we construct a reasoning-focused calibration set from a small pre-specified grid: $n \in \{128, 256\}$ examples per task and $K \in \{3,5\}$ subsamples, all with fixed random seeds ($\text{seed}=0$).
All models are decoded with greedy sampling (temperature $= 0$) using chain-of-thought prompts adapted from prior work \citep{Wei2022ChainOfThought}.
We report results at 10\%, 20\%, and 50\% global sparsity.

\paragraph{Baseline selection.}
CAP is compared against the original dense (unpruned) model and Wanda \citep{sun2024simpleeffectivepruningapproach}.
We select Wanda as the primary baseline because it operates in the same computational regime as CAP: both are training-free, one-shot, require no weight updates, and use the same calibration data for computing pruning scores.
This makes Wanda the fairest like-for-like comparison.
SparseGPT \citep{frantar2023sparsegptmassivelanguagemodels} represents a stronger but more expensive alternative that applies compensatory weight updates via second-order information during pruning; we discuss the relationship between CAP and SparseGPT, and the implications of this baseline scope, in Section~\ref{sec:limitations}.

\paragraph{Evaluation metrics.}
We measure reasoning preservation through \emph{final-answer accuracy}: whether the model produces the correct answer after chain-of-thought generation.
This is a standard and widely used metric for reasoning benchmarks, but it does not directly assess the quality of intermediate reasoning steps---a pruned model might reach correct answers through degraded reasoning paths, or produce sound reasoning but fail at the answer-extraction step.
We discuss this evaluation scope further in Section~\ref{sec:eval_scope}.

\section{Results}

We evaluate CAP against the Wanda baseline on three reasoning benchmarks.
Table~\ref{tab:llama-mistral} summarizes results for Llama-3-8B-Instruct and Mistral-7B-Instruct, including perplexity on WikiText-2 as a complementary measure of general language modeling quality.

\subsection{Main Findings: Moderate Sparsity (10--20\%)}

At 10\% and 20\% sparsity, CAP usually preserves reasoning-benchmark performance better than Wanda, though the gains vary by model and benchmark.
The largest improvements appear on ARC-Challenge for Llama-3.

On Llama-3-8B-Instruct at 10\% sparsity, CAP achieves 71.5\% accuracy on GSM8K compared to Wanda's 59.5\%, a 20\% relative improvement.
For StrategyQA, CAP maintains 65.5\% versus Wanda's 65.0\%, and on ARC-Challenge achieves 74.2\% versus 68.7\%.
At 20\% sparsity, CAP's advantage becomes even more striking on ARC-Challenge: 70.8\% accuracy versus Wanda's 43.9\%, a 61\% relative improvement.
This benefit is not universal, however: on Llama-3 GSM8K at 20\% sparsity, CAP trails Wanda (65.4\% vs.\ 72.1\%).
One plausible explanation is that under heavier pruning, CAP's head-level protection may better preserve benchmark performance on multi-step inference tasks than the precise answer-formatting behavior needed for GSM8K final-answer extraction.
The ARC-Challenge gap is at least consistent with the hypothesis that Wanda's correlational criterion harms behaviors useful for scientific reasoning and knowledge integration more than CAP's task-aligned scoring.

On Mistral-7B-Instruct, the pattern is more nuanced.
At 10\% sparsity, both methods perform comparably, with differences within 1.5 percentage points.
At 20\% sparsity, CAP maintains slight advantages on StrategyQA and GSM8K but trails Wanda on ARC-Challenge.
The smaller margins on Mistral may reflect its more diffuse head importance distribution (Appendix~\ref{app:head_heatmaps}), which provides less separation for CAP to exploit.

By explicitly measuring $\Delta(h_\ell)$ via interventional masking, CAP can distinguish between heads that are numerically small but functionally critical and heads that are large but redundant---a distinction that activation-based proxies cannot reliably make.

\subsection{High Sparsity (50\%): Limitations}
\label{sec:high_sparsity}

At 50\% sparsity, both methods suffer severe degradation, though they fail differently.
On Llama-3, Wanda maintains modest GSM8K and StrategyQA accuracy while CAP drops substantially on these tasks.
However, CAP preserves better ARC-Challenge performance.
The perplexity divergence is stark: Wanda maintains 55.6 while CAP's perplexity increases to 428.2, indicating collapse of general language modeling.

This failure stems from CAP's coarse-grained treatment of MLP layers---assigning a single layer-average importance derived from attention heads.
When the global threshold becomes aggressive at 50\% sparsity, systematically important MLP weights are assigned inappropriately low importance, leading to their removal and model collapse.

In contrast, on Mistral at 50\% sparsity, CAP outperforms Wanda across all reasoning tasks and maintains reasonable perplexity (10.5 vs.\ 9.8), which may indicate that Mistral is more amenable to aggressive causal-guided pruning.

\subsection{Perplexity vs.\ Reasoning Performance}

At 10--20\% sparsity, CAP often achieves better reasoning accuracy than Wanda while exhibiting slightly higher perplexity.
On Llama-3 at 20\% sparsity, CAP shows perplexity of 9.6 versus Wanda's 8.7, yet substantially outperforms Wanda on ARC-Challenge (70.8\% vs.\ 43.9\%) and StrategyQA (63.1\% vs.\ 60.5\%).
This decoupling suggests that CAP's reasoning-focused calibration can preserve reasoning-benchmark performance even when general token prediction degrades somewhat---a trade-off that may be desirable when reasoning is the deployment priority.

\subsection{Preliminary Observation on Mixture-of-Experts Architectures}

In preliminary exploratory experiments on a mixture-of-experts (MoE) model, CAP appeared to underperform Wanda across the sparsity levels we tested.
We treat this as a qualitative limitation rather than a benchmarked empirical finding, since we do not report model-specific numbers here.
The sparse expert routing in MoE models introduces a fundamental challenge for head-level causal attribution: routing decisions are dynamically determined per token, so a head's importance is not stationary.
Our calibration procedure computes a single $\Delta(h_\ell)$ score per head by averaging across the calibration set, but in an MoE model the same head may be critical for some expert combinations and irrelevant for others.
This suggests that CAP in its current form is less suitable for MoE architectures and likely requires routing-aware extensions.

\subsection{Summary}

CAP is most effective at preserving reasoning-benchmark performance under moderate compression (10--20\% sparsity) by explicitly measuring and protecting causally important attention heads.
It outperforms Wanda in most model--benchmark settings within this range, with individual gains reaching 61\%.
However, CAP exhibits limitations at high sparsity ($\geq$50\%) where coarse MLP treatment can cause catastrophic failure, and our preliminary MoE experiments suggest weaker transfer to dynamically routed architectures.


\begin{table*}[t]
\centering
\caption{Evaluation on Llama-3-8B-Instruct and Mistral-7B-Instruct ($n{=}128$ samples per task, $K{=}3$ calibration subsets, median aggregation). Bold indicates the best pruning method per sparsity for each \emph{model $\times$ benchmark}. Perplexity (Perp.): lower is better; accuracy: higher is better.}\label{tab:llama-mistral}
\setlength{\tabcolsep}{3.5pt}
\small
\begin{tabular}{l c | cccc | cccc}
\toprule
 &  & \multicolumn{4}{c|}{\textbf{Llama-3-8B-Instruct}} & \multicolumn{4}{c}{\textbf{Mistral-7B-Instruct}} \\
\textbf{Method} & \textbf{Sparsity} & \textbf{GSM8K} & \textbf{StrategyQA} & \textbf{ARC-C.} & \textbf{Perp.} & \textbf{GSM8K} & \textbf{StrategyQA} & \textbf{ARC-C.} & \textbf{Perp.} \\
\midrule
\textbf{Dense} & 0\%
& 79.5 & 69.5 & 81.1 & 6.23
& 42.9 & 64.3 & 55.2 & 6.20 \\
\midrule
Wanda & 10\%
& 59.5 & 65.0 & 68.7 & \textbf{8.37}
& \textbf{46.4} & 65.1 & 34.0 & \textbf{5.83} \\
\textbf{CAP} & 10\%
& \textbf{71.5} & \textbf{65.5} & \textbf{74.2} & 8.44
& 45.7 & \textbf{65.3} & \textbf{34.2} & 5.85 \\
\midrule
Wanda & 20\%
& \textbf{72.1} & 60.5 & 43.9 & \textbf{8.70}
& 43.6 & 63.8 & \textbf{34.9} & \textbf{5.89} \\
\textbf{CAP} & 20\%
& 65.4 & \textbf{63.1} & \textbf{70.8} & 9.57
& \textbf{44.4} & \textbf{64.8} & 33.4 & 5.93 \\
\midrule
Wanda & 50\%
& \textbf{1.7} & \textbf{51.7} & 8.5 & \textbf{55.6}
& 11.5 & 49.5 & 44.7 & \textbf{9.82} \\
\textbf{CAP} & 50\%
& 0.5 & 6.9 & \textbf{30.1} & 428.2
& \textbf{27.3} & \textbf{53.0} & \textbf{52.6} & 10.5 \\
\bottomrule
\end{tabular}
\end{table*}

\section{Analysis and Ablations}
\label{sec:ablation}

We report targeted qualitative analyses from development runs to clarify which design choices appear to drive CAP's performance and where the method is most likely to fail.
Unless otherwise stated, these observations are intended as directional support rather than exhaustive quantitative ablations.

\paragraph{Effect of reasoning-focused calibration.}
We compare CAP calibrated on our mixed GSM8K/StrategyQA/ARC-Challenge set with CAP calibrated on WikiText-2 samples of comparable size.
When calibrated on WikiText-2, CAP achieves similar perplexity but tends to underperform on GSM8K and ARC-Challenge, particularly at 20\% sparsity.
This supports the hypothesis that calibration distribution should match the target capability.

\paragraph{Median vs.\ mean aggregation.}
Replacing the median with a mean for head-importance aggregation tends to produce less stable rankings across random seeds and calibration splits, manifesting as higher variance in downstream accuracy and occasional large drops on ARC-Challenge.
Median aggregation reduces the influence of outlier examples where masking a head produces unusually large loss spikes.

\paragraph{Weight-level vs.\ head-level pruning.}
Comparing CAP's weight-level pruning with a structural variant that removes entire heads based on causal scores reveals that structural pruning is brittle: once a small number of high-importance heads are mistakenly removed, both perplexity and reasoning accuracy degrade sharply.
Weight-level pruning allows CAP to preferentially remove parameters from low-importance heads while retaining some capacity in those regions, and to prune only the numerically smallest weights within critical heads.

\paragraph{Number of calibration subsamples $K$.}
Moving from $K=1$ to $K=3$ materially stabilizes head rankings and reduces accuracy variance in development runs.
$K=5$ provides smaller additional gains at higher compute cost.
We therefore restrict the main experiments to the small range $K\in\{3,5\}$.

\paragraph{Architecture-specific behavior.}
Analysis of causal head importance patterns (Appendix~\ref{app:head_heatmaps}) reveals that Llama-3 concentrates critical heads in a narrow band of layers, whereas Mistral spreads importance more uniformly.
CAP is particularly effective when there is a clear separation between high- and low-importance regions, as in Llama-3 at moderate sparsity.

\subsection{Scope of Reasoning Evaluation}
\label{sec:eval_scope}

Our evaluation measures reasoning preservation through final-answer accuracy: whether the pruned model produces the correct answer after chain-of-thought generation.
This is a widely used metric that captures the model's end-to-end reasoning capability, but it has important limitations that we acknowledge.

Final-answer accuracy does not directly assess the quality of intermediate reasoning steps.
A pruned model might arrive at a correct answer through a degraded or shortcut reasoning path, or conversely produce a coherent reasoning chain but fail at the final extraction step.
A richer evaluation would analyze intermediate reasoning along several dimensions:
\emph{chain completeness} (whether the model produces all necessary reasoning steps without truncation),
\emph{logical coherence} (whether each step follows from the previous one),
\emph{error categorization} (distinguishing arithmetic errors from logical errors or incoherent outputs), and
\emph{step count preservation} (whether pruning reduces the number of reasoning steps the model generates).

While such analyses are feasible in principle, a thorough intermediate reasoning evaluation across all models, sparsity levels, and benchmarks was beyond our computational budget for this study.
We report final-answer accuracy as a necessary but acknowledged incomplete measure of reasoning preservation, and we consider more fine-grained reasoning analysis an important direction for future work.

\section{Limitations and Future Work}
\label{sec:limitations}

We highlight several limitations of the current work.

\paragraph{Baseline scope.}
Our experimental comparison centers on Wanda.
Broader comparisons against methods such as SparseGPT \citep{frantar2023sparsegptmassivelanguagemodels}, which applies compensatory weight updates, and simple magnitude pruning as a lower bound would strengthen the empirical picture, especially at high sparsity.

\paragraph{Evaluation granularity.}
As discussed in Section~\ref{sec:eval_scope}, we evaluate reasoning preservation only through final-answer accuracy, which does not capture intermediate reasoning quality.
Systematic analysis of chain-of-thought coherence, step completeness, and error modes would provide a richer understanding of how pruning affects reasoning processes.

\paragraph{Scope of causal attribution.}
CAP's causal scoring is defined over attention heads only.
MLP layers---which store factual knowledge and implement key computations---receive a coarse per-layer importance average, which becomes increasingly inadequate at high sparsity.
This mismatch is the primary cause of CAP's catastrophic failure on Llama-3 at 50\% sparsity.
Extending causal interventions to MLP neurons (e.g., via neuron-level activation patching) would address this gap, though at significant computational cost.

\paragraph{High sparsity.}
At sparsity levels above ${\sim}$40\%, CAP's global thresholding can remove parameters even within causally critical heads, and the coarse MLP treatment exacerbates collapse. CAP is best suited for moderate sparsity (10--30\%).

\paragraph{Architecture generalization.}
Our preliminary exploratory tests suggest that CAP does not transfer cleanly to mixture-of-experts architectures because head importance becomes non-stationary under dynamic routing. Routing-aware extensions are needed.

\paragraph{Inference efficiency.}
While CAP targets unstructured weight sparsity, hardware and software support for unstructured sparsity is still maturing. Advancing sparse inference infrastructure remains important for realizing the latency benefits of unstructured pruning.

\paragraph{Perplexity--reasoning trade-off.}
At moderate sparsity, CAP often achieves better reasoning performance but slightly worse perplexity, suggesting a trade-off between general language modeling and task-specific capability preservation. Understanding this decoupling is an interesting direction for future investigation.

\section{Conclusion}

We introduced Causal Attribution Pruning (CAP), a training-free method for preserving reasoning-benchmark performance under moderate sparsity in compressed language models.
CAP measures the causal contribution of individual attention heads to task performance and uses these measurements to guide weight-level pruning.
At moderate sparsity (10--20\%), CAP outperforms Wanda in most model--benchmark settings, with relative gains of up to 61\% on ARC-Challenge, suggesting that interventional, task-aligned pruning of attention heads can better preserve performance on reasoning benchmarks than correlational proxies.

Our work also reveals important limitations and open questions.
CAP's causal scoring is defined over attention heads and does not extend to fine-grained MLP attribution, leading to failure at high sparsity on certain architectures.
Our evaluation relies on final-answer accuracy, which does not directly assess intermediate reasoning quality.
Our comparison is primarily against a single baseline (Wanda), and extending to methods like SparseGPT would provide a more complete picture.

These limitations suggest several directions for future work: neuron-level causal attribution for MLP layers, intermediate reasoning quality metrics for pruning evaluation, broader baseline comparisons, and extensions to dynamic architectures such as mixture-of-experts models.
More broadly, our results underscore the value of task-aware, interventional pruning objectives when the goal is to preserve specific capabilities rather than general language modeling quality.

\section*{Ethical Statement}

This work studies pruning methods for existing open-weight language models. We do not train new models from scratch, and we limit our experiments to widely used research benchmarks (GSM8K, StrategyQA, ARC-Challenge, WikiText-2) that do not contain personally identifiable or sensitive information. All models are used in accordance with their respective licenses.

Model compression can have both positive and negative societal impacts. Improved pruning can lower the computational and energy costs of deploying language models, reducing environmental impact and making advanced models more accessible. However, lowering deployment costs may also facilitate misuse. Our work does not directly address these risks, and we encourage future research to consider how compression interacts with safety interventions.

No human subjects, crowd workers, or proprietary user data were involved in this study.

\section*{Acknowledgments}
We thank Algoverse for supporting this research.

\bibliographystyle{iclr2026_conference}
\bibliography{references}

\appendix
\section{Causal Head Importance Visualizations}
\label{app:head_heatmaps}

\begin{figure}[h]
\centering
\begin{subfigure}[b]{0.48\linewidth}
  \centering
  \includegraphics[width=0.9\linewidth]{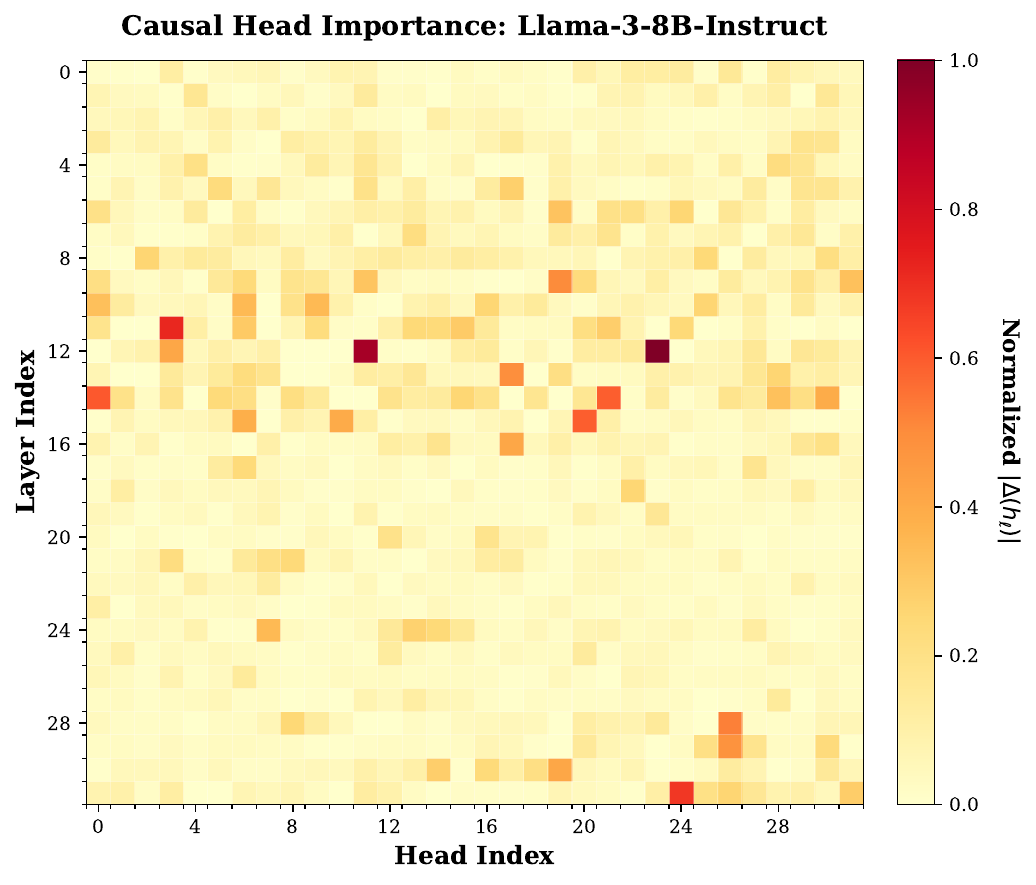}
  \caption{Llama-3-8B-Instruct}
\end{subfigure}
\hfill
\begin{subfigure}[b]{0.48\linewidth}
  \centering
  \includegraphics[width=0.9\linewidth]{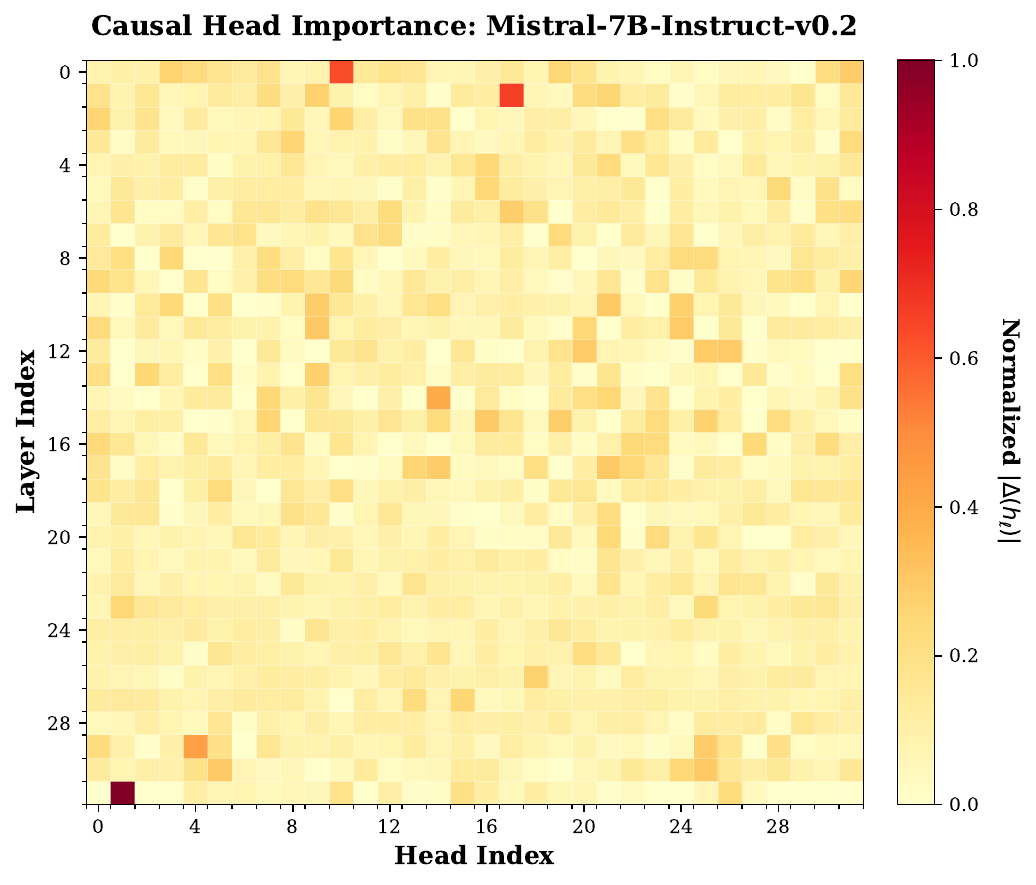}
  \caption{Mistral-7B-Instruct-v0.2}
\end{subfigure}
\caption{\textbf{Causal head importance across layers.} Each cell $(i,j)$ shows the normalized $|\Delta(h_\ell)|$ for head $j$ in layer $i$, computed on the reasoning calibration set (128 samples each from GSM8K, StrategyQA, ARC-Challenge). Darker red indicates higher causal importance. \textbf{Llama-3} exhibits concentrated critical heads in layers 11--14 and at layer boundaries, with substantial redundancy in layers 15--27. \textbf{Mistral} shows more diffuse importance with critical heads in early layers (0--5). Both models have $<$20\% of heads with high causal scores, validating the opportunity for aggressive pruning guided by interventional measurements.}
\label{fig:head_heatmap}
\end{figure}

Figure~\ref{fig:head_heatmap} visualizes the causal head importance patterns across layers for both models.
These heatmaps reveal the variability in head utility that motivates CAP's weight-level pruning approach.

\section{Evaluation Prompt Templates}
\label{app:prompts}

All evaluations use deterministic greedy decoding (temperature $=0$) with chain-of-thought prompting.
Prompts instruct the model to reason step-by-step and output the final answer in a strict format.

\subsection{GSM8K}

\begin{listing}[h]
\caption{GSM8K prompt template}
\begin{lstlisting}
prompt = f"""{q}
Please solve the problem step-by-step.
Then, write your final answer on the last
line with the numerical value only in the
format:
Answer: <number>"""
\end{lstlisting}
\end{listing}

\noindent
The numeric answer is extracted from the gold solution following the pattern \texttt{\#\#\#\# <number>}.

\subsection{StrategyQA}

\begin{listing}[h]
\caption{StrategyQA prompt template}
\begin{lstlisting}
prompt = f"""{q}
Please reason step-by-step. Then, on the
last line, output only:
Answer: Yes
or
Answer: No"""
\end{lstlisting}
\end{listing}

\subsection{ARC-Challenge}

\begin{listing}[!htbp]
\caption{ARC-Challenge prompt template}
\begin{lstlisting}
prompt = f"""{q}
{opts_block}
Please solve step-by-step. Then, on the
last line, output only:
Answer: <letter>"""
\end{lstlisting}
\end{listing}

\noindent
where \texttt{\{opts\_block\}} contains the formatted multiple-choice options (e.g., \texttt{A.~first choice}, \texttt{B.~second choice}, etc.).

\vspace{0.5em}
\noindent
Across all tasks, the model produces a step-by-step reasoning trace, but evaluation extracts \textbf{only the final line} matching the answer format.

\end{document}